%% file: main.tex
\documentclass[letterpaper, 10 pt, conference]{ieeeconf}  

\IEEEoverridecommandlockouts                              

\usepackage{amsmath,amssymb,amsfonts}
\usepackage{bm} 
\usepackage{algorithm}
\usepackage{algpseudocode} 
\usepackage{graphicx}
\usepackage{caption}
\usepackage{subcaption}   
\usepackage{dblfloatfix}  
\usepackage{xcolor}
\usepackage{graphicx}
\usepackage{subcaption}
\usepackage{booktabs}
\usepackage{placeins}   
\usepackage{rotating}   
\usepackage{float}
\usepackage{placeins}  
\usepackage{graphicx}
\usepackage{xcolor}
\usepackage[font=footnotesize,labelfont=bf]{caption}

\title{\LARGE \bf
Surface-Constrained Offline Warping with Contact-Aware Online Pose Projection for Safe Robotic Trajectory Execution
}

\author{
Farong Wang$^{1}$,
Sai Swaminathan$^{1}$,
Fei Liu$^{1}$
\thanks{This work was not supported by any organization.}
\thanks{$^{1}$All authors are with the Min H. Kao Department of Electrical Engineering and Computer Science (EECS),
University of Tennessee, Knoxville (UTK), Knoxville, TN 37996, USA.}
\thanks{\texttt{\small \{fwang31, sai, feiliu\}@utk.edu}}
\thanks{We thank undergraduate students in UTK EECS for assistance with system implementation and evaluation:
Benjamin Drumwright, Trevor Olson (Blender and trajectory visualization), and Begad Almany (deformable demo building).}
}

\begin{document}

\maketitle
\thispagestyle{empty}
\pagestyle{empty}

\begin{abstract}
Robotic manipulation tasks that require repeated tool motion along curved surfaces frequently arise in surface finishing, inspection, and guided interaction. In practice, nominal motion primitives are often designed independently of the deployment surface and later reused across varying geometries. Directly tiling such primitives onto nonplanar surfaces introduces geometric inconsistencies, leading to interpenetration, orientation discontinuities, and cumulative drift over repeated cycles. We present a two-stage framework that separates geometric embedding from execution-level regulation. An offline surface-constrained warping operator embeds a nominal periodic primitive onto curved surfaces through asymmetric diffeomorphic deformation of dual-track waypoints and axis-consistent orientation completion, producing a surface-adapted reference trajectory. An online contact-aware projection operator then enforces bounded deviation relative to this reference using FSR-driven disturbance adaptation and a conic orientation safety constraint. Experiments across multiple analytic surface families and real-robot validation on a sinusoidal surface demonstrate improved geometric continuity, reduced large orientation jumps, and robust contact maintenance compared with direct tiling. These results show that decoupling offline geometric remapping from lightweight online projection enables stable and repeatable surface-embedded trajectory execution under sensor-lite feedbacks.
\end{abstract}

\input{introduction}
\input{method}

\begin{figure}[!h]
  \centering
  \includegraphics[width=\linewidth]{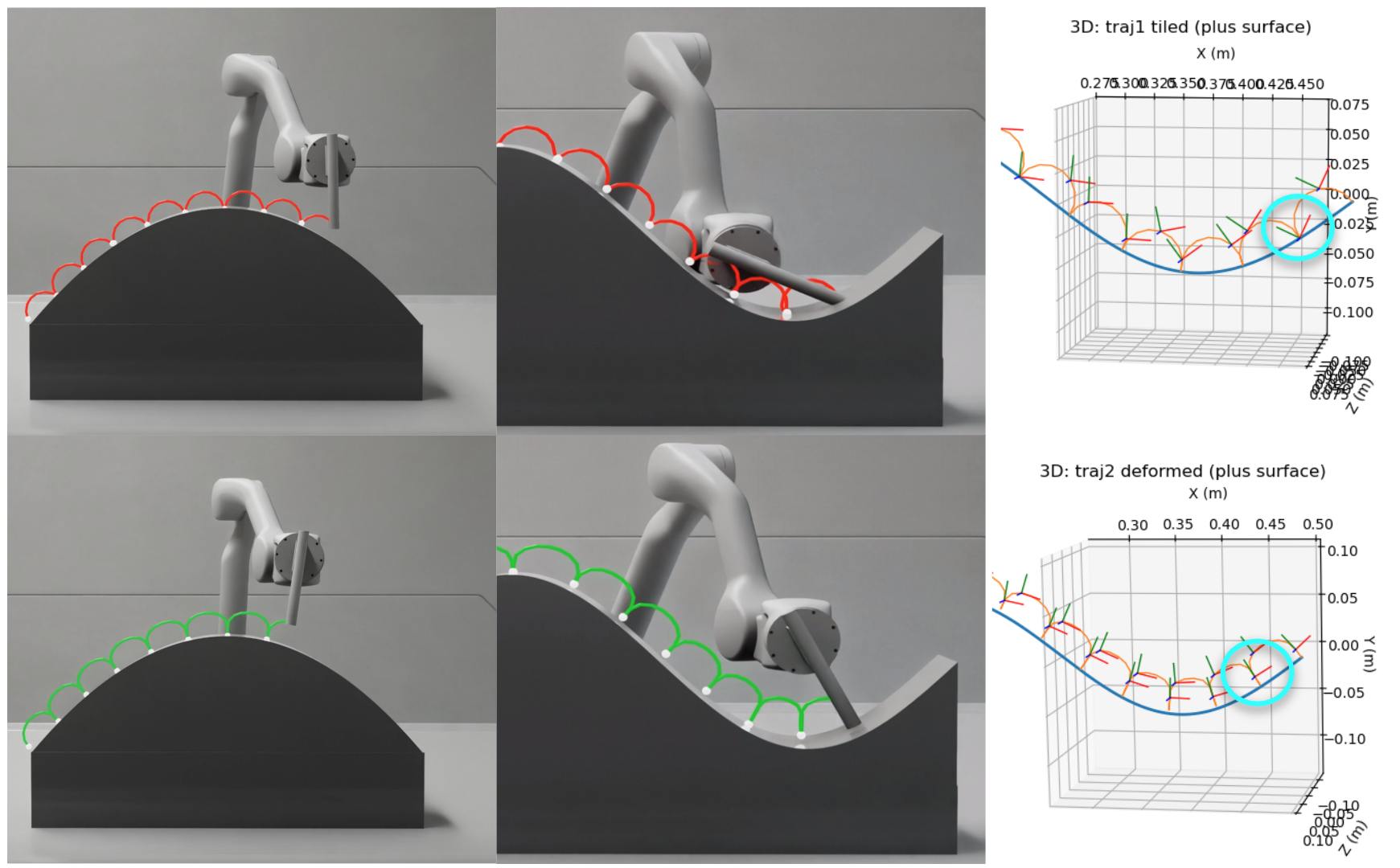}
  \caption{\textbf{Collision-prone configurations under direct tiling.}
First row: \texttt{tiled} (\(\mathbf T^{\mathrm{tile}}\)). 
Second row: \texttt{warped} (\(\mathbf T^{\mathrm{warp}}\)).
Direct tiling remains feasible on the convex surface but becomes collision-prone on the concave surface due to an unsafe local attitude. Offline warping produces a surface-consistent trajectory that avoids this failure mode.}
  \label{fig:offline_collision}
\end{figure}

\begin{figure}[!h]
  \centering
  \includegraphics[width=\linewidth]{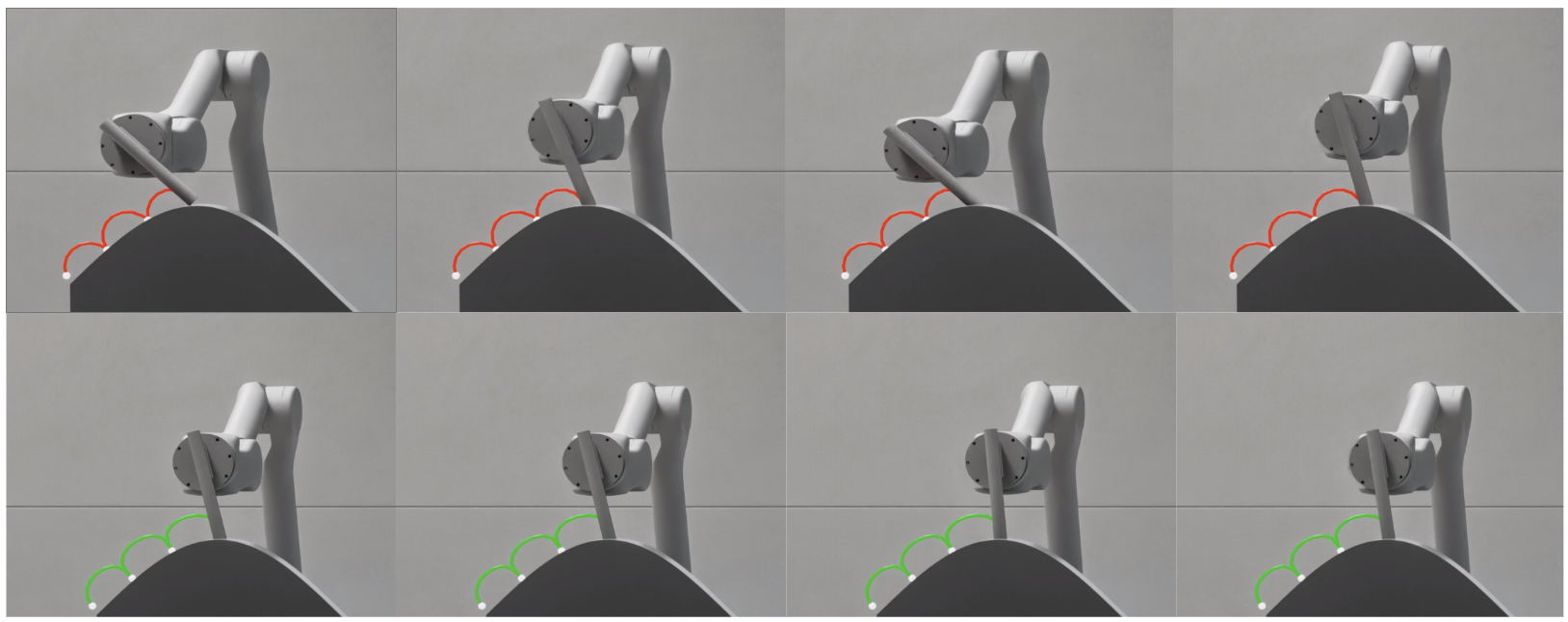}
  \caption{\textbf{Continuity correction on a convex surface.}
  First row: \texttt{tiled}. Second row: \texttt{warped}.
  Compared with direct tiling, the warped trajectory yields smoother and more monotone local pose evolution along the same guide neighborhood.}
  \label{fig:offline_cont}
\end{figure}

\begin{figure*}[!ht]
  \centering
  \includegraphics[width=\textwidth]{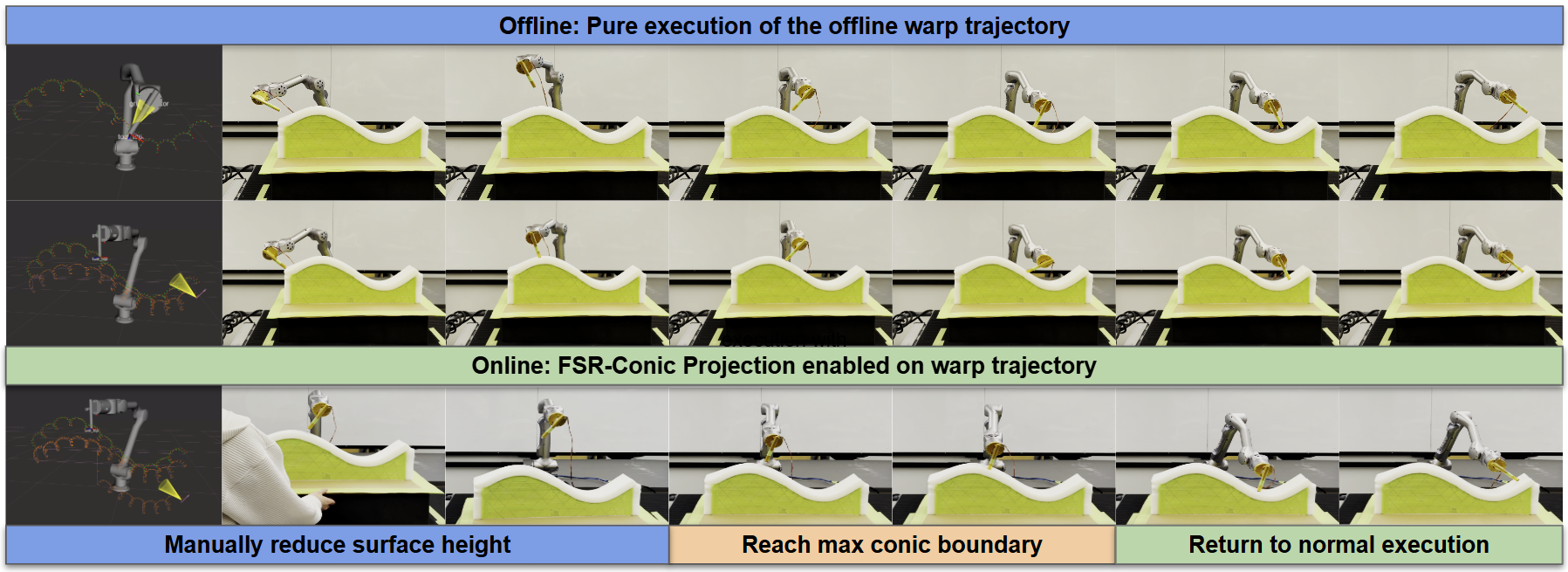}
  \caption{\textbf{Real-robot validation under disturbances and abrupt height reduction.}
The proposed FSR-conic Projection improves contact consistency during disturbed execution and recovers from sudden surface-height reduction through translational adaptation, while conic filtering maintains bounded orientation deviation throughout the motion.}
  \label{fig:robot_set1}
\end{figure*}

\section{Experiments}
\label{sec:experiments}

We evaluate the proposed two-stage framework from two perspectives. Offline, we verify that $W_\Phi$ resolves the main failure modes of direct periodic tiling and produces surface-consistent warped trajectories. Online, we validate that $E$ improves contact robustness on a real robot under designed disturbances and abrupt environment change.

\subsection{Offline Geometric Validation}
\label{sec:exp_offline}

We compare direct tiling ($T^{\mathrm{tile}}$) with offline warping ($T^{\mathrm{warp}}$). On non-planar surfaces, direct tiling exhibits two main failure modes that limit reliable execution: collision-prone attitudes in concave regions and non-smooth local pose evolution along the guide. The offline operator $W_\Phi$ reshapes the tiled sequence into a surface-consistent warped trajectory that addresses both.

Figure~\ref{fig:offline_collision} illustrates the first failure mode. In the convex case (left), direct tiling remains feasible and follows the intended guide without obvious geometric conflict. In the concave case (right), however, the same tiling strategy produces an unsafe attitude that collides with the local surface, even though the tool tip remains close to the desired guide. The second row shows the corresponding \texttt{warped} trajectory. After offline warping, the pose sequence becomes surface-consistent and avoids this unsafe configuration.

The second failure mode is local discontinuity. Even when no collision occurs, \texttt{tiled} can exhibit reciprocal or back-and-forth pose changes within the same guide neighborhood, leading to non-monotone orientation evolution. Figure~\ref{fig:offline_cont} compares \texttt{tiled} and \texttt{warped} on a convex surface. Direct tiling produces irregular local pose progression, whereas offline warping yields a smoother and more monotone sequence while preserving the intended periodic embedding.

To quantify this improvement, we perform a parameter-sweep evaluation over five analytic surface families (\texttt{sin}/\texttt{cos}/\texttt{exp}/\texttt{parabolic}/\texttt{cubic}) using paired \texttt{tiled}/\texttt{warped} trajectories. For each trajectory, we compute the per-step angular change
\(\Delta \theta_k=d(q_{k+1},q_k)\),
and report two continuity metrics: the 95th-percentile angular step
\(\mathrm d\theta_{p95}\) and the bad-step rate
\(\mathbb P(\Delta \theta_k>10^\circ)\),
which captures the large rotational jumps responsible for local oscillation. Table~\ref{tab:smoothness_by_surface} reports paired trajectory-level aggregates. Across all surface families, \texttt{warped} reduces large rotation-step outliers, with the strongest improvement on higher-curvature families such as \texttt{sin} and \texttt{cos}. On \texttt{cubic}, the gain is marginal because direct tiling already exhibits near-zero discontinuity.

\begin{table}[!h]
\centering
\caption{\textbf{Surface-family continuity summary (warped vs.\ tiled).}
\#Pair denotes the number of paired tiled/warped trajectories, and \#Step denotes the total number of evaluated pose transitions from the tiled trajectories.
Median \(\Delta(\mathrm{d}\theta_{p95})\) (deg) is \texttt{warped} minus \texttt{tiled} (negative is better).
BadRot$_t$ / BadRot$_w$ are the average fractions of steps with \(\Delta \theta_k > 10^\circ\) for \texttt{tiled}/\texttt{warped}.}
\label{tab:smoothness_by_surface}
\setlength{\tabcolsep}{3.5pt}
\renewcommand{\arraystretch}{1.05}
\begin{tabular}{l r r r r r r}
\toprule
Surf. & \#Pair & \#Step & Med.\,$\Delta(\mathrm{d}\theta_{p95})$ & BadRot$_t$ & BadRot$_w$ & $\Delta$ \\
\midrule
cos       & 4 & 463  & -14.98 & 0.093 & 0.000 & 0.093 \\
cubic     & 1 & 98   & -0.05  & 0.000 & 0.000 & 0.000 \\
exp       & 3 & 1121 & -0.99  & 0.005 & 0.000 & 0.005 \\
parabolic & 4 & 436  & -6.77  & 0.037 & 0.000 & 0.037 \\
sin       & 4 & 471  & -16.59 & 0.125 & 0.016 & 0.109 \\
\bottomrule
\end{tabular}
\end{table}

Overall, the offline results show that \(\mathcal W_{\Phi}\) removes the two main geometric blockers of direct tiling---collision-prone attitudes and oscillatory discontinuities---while substantially improving orientation continuity across diverse surfaces.


\subsection{Online Real-Robot Validation}
\label{sec:exp_online_robot}

We validate the online operator \(\mathcal E\) directly on hardware, since its purpose is to improve contact robustness using minimal sensing during execution. We use a representative sinusoidal surface, whose rapidly varying local normal amplifies tracking error and makes open-loop warp execution particularly challenging.

Figure~\ref{fig:robot_set1} summarizes the real-robot results. The top row shows pure execution of the offline warp trajectory without online correction. Although the warp trajectory itself is geometrically feasible, the resulting contact quality is uneven, with especially large mismatch around the \(\pi/2\) region and only partial agreement in other intended contact segments. The middle row shows execution of the same warp trajectory with the proposed online operator enabled. Under intentionally injected FSR-coupled disturbances, the robot maintains contact more consistently across the intended contact segments, indicating that online disturbance correction improves local adaptation while the conic filter bounds orientation deviation around the warp trajectory.

The bottom row of Fig.~\ref{fig:robot_set1} evaluates robustness under abrupt height reduction. During execution, we remove the support base to lower the surface unexpectedly. The robot recovers contact through translational adaptation, while the conic filter prevents excessive rotational drift even under large disturbance. This result shows that the online operator can tolerate sudden environment change without losing bounded-orientation safety.

Taken together, the real-robot results show that \(\mathcal E\) improves contact robustness using only a 1D FSR signal, while the warp-centered conic filter provides non-drifting orientation safety under both designed disturbances and abrupt environment change.

\section*{Conclusion}
\begin{figure}[!h]
  \centering
  \includegraphics[width=\linewidth]{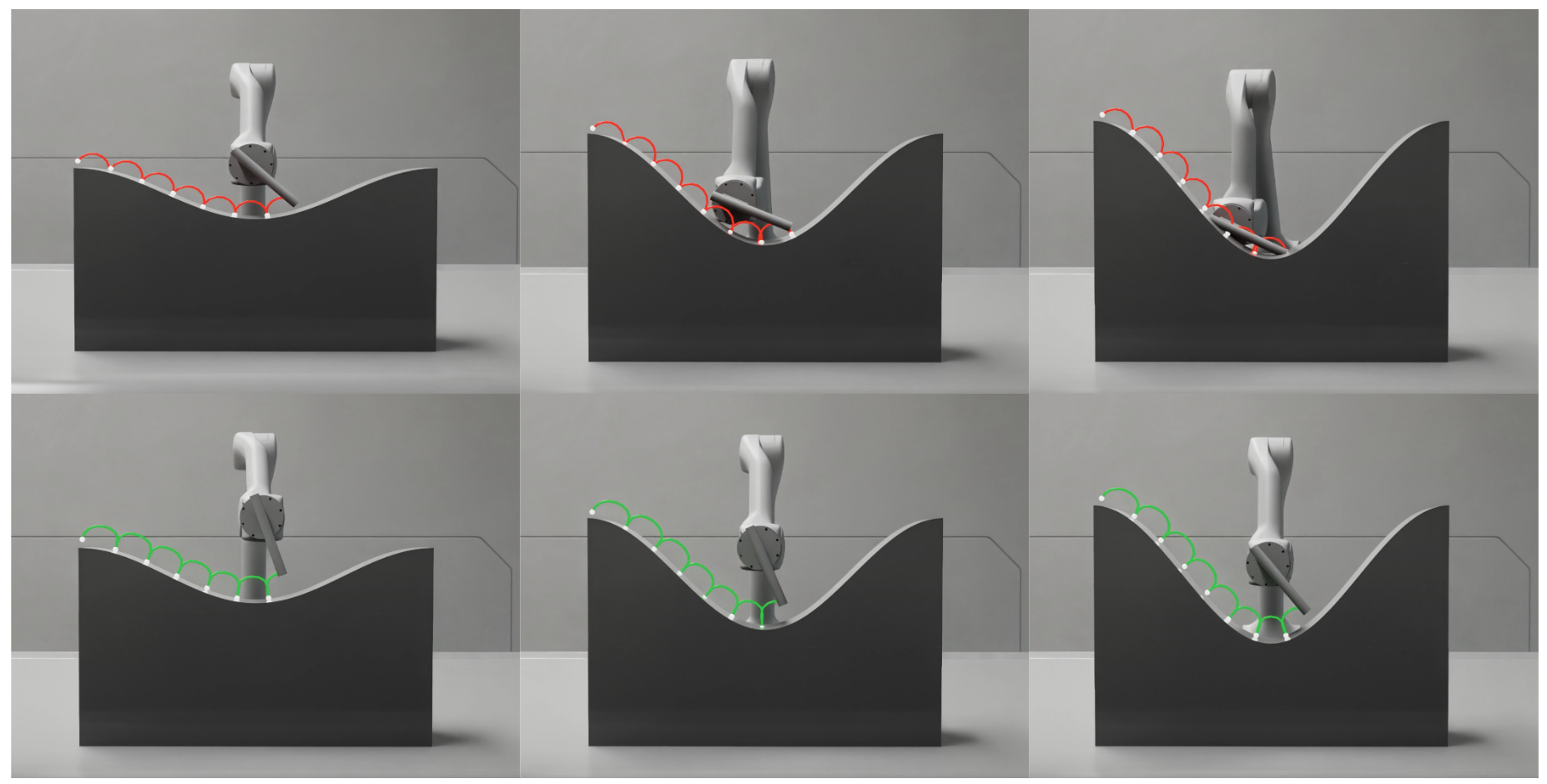}
  \caption{\textbf{Representative simulation results of offline warping.}
Examples compare tiled ($T^{\mathrm{tile}}$) and warped ($T^{\mathrm{warp}}$) trajectories. Direct tiling may produce collision-prone or geometrically inconsistent poses near complex surface regions, whereas offline warping yields surface-consistent and smoother motion.}
  \label{fig:offline_collision_end}
\end{figure}

We presented a two-stage framework for executing periodic end-effector skills on curved surfaces with continuous pose evolution and bounded orientation deviation under contact. Offline, the warping operator \(\mathcal W_{\Phi}\) resolves collision-prone attitudes and local oscillatory orientation changes caused by direct tiling on non-planar surfaces. Online, the execution operator \(\mathcal E\) improves contact robustness through bounded local correction, while a warp-centered conic filter limits orientation drift relative to the warped reference. Across 32 trajectories over five analytic surface families and real-robot experiments on a sinusoidal surface, the results support the effectiveness of the proposed framework. Figure~\ref{fig:offline_collision_end} further illustrates representative offline warping results across different surface geometries.

\bibliographystyle{IEEEtran}
\bibliography{references}
\end{document}

%% file: introduction.tex
\section{Introduction}
Robotic tasks that require a tool to move along a curved surface while maintaining consistent contact arise in surface finishing, inspection, and surface-guided manipulation \cite{Rui_Wu_2025_IROS, Schneyer_2023, Fengjie_Tian_2016}. In many practical workflows, a nominal motion primitive, often periodic, is first designed in a canonical frame and later reused across different surface regions or workpieces \cite{Saveriano_2023_IJRR, Christoph_2024_IROS, Liang_Han_2023}. Directly replaying or tiling such primitives on nonplanar geometries can introduce geometric inconsistencies. Even small mismatches between the primitive and local surface curvature accumulate over repeated cycles. This may result in interpenetration, discontinuous orientation changes, and progressive drift. The problem is fundamentally geometric. It concerns how to transfer structured motion defined in one geometric domain into another while preserving continuity and internal structure.

\begin{figure}[!t]
    \centering
    \includegraphics[width=0.45\textwidth]{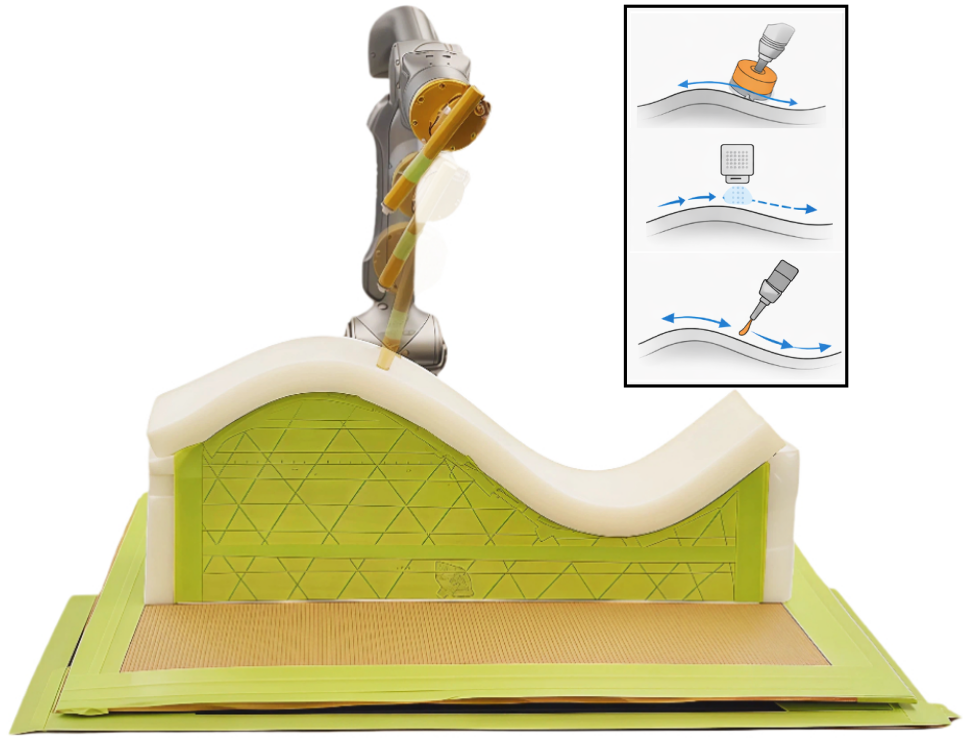}
    \caption{\textbf{Surface-constrained robotic trajectory execution with stable contact.}
    A robotic end-effector follows a warped periodic trajectory over a curved surface using offline geometric warping and online contact-aware projection for stable and drift-free interaction, with applications in polishing, coating, inspection, and surgical assistance.
    }
    \label{fig:cover_picture}
\end{figure}

\subsection{Related Works}

Prior work has addressed surface interaction from two complementary perspectives. The first focuses on geometric path generation or tracking directly on the target surface. Surface-aware tool path generation methods compute coverage-consistent trajectories that respect intrinsic surface geometry \cite{Yalun_2022_RAL}. Curved surface path following has also been formulated as a constrained dynamical tracking problem, where model predictive controllers enforce surface consistency during execution \cite{Santos_2023_RAL}. These approaches assume that the reference trajectory is already surface-consistent and concentrate on online optimization or tracking.

A second line of research embeds surface constraints directly into motion representations. Manifold-aware learning and planning methods generate trajectories intrinsically on geometric constraint manifolds. Duan et al.~\cite{Anqing_2025_TRO} learn rhythmic trajectories under explicit Riemannian constraints, enabling motion generation directly on surface manifolds. Mesh-based movement primitives encode trajectories intrinsically on triangulated surfaces using discrete differential geometry operators and geodesic computations \cite{Vedove_2025}. Constrained planning methods further generate continuous paths that remain on learned or specified manifolds \cite{Canzini_2024}. These approaches incorporate geometric structure during learning or synthesis. However, they typically assume that demonstrations or policies are already defined intrinsically on the surface.

Despite these advances, an important practical scenario remains insufficiently addressed. In many manipulation workflows, nominal motion primitives are designed or learned in canonical task frames and later deployed across workpieces with varying surface geometries. Movement primitive representations and dynamical skill models are often constructed independently of a specific deployment surface and subsequently adapted to new constraints or contexts \cite{Saveriano_2023_IJRR, Frank_2022_TRO, Davchev_2022_RAL, Thibaut_2020}. Geometry-aware tracking research further emphasizes that geometric structure must be explicitly considered to maintain consistency and manipulability during motion transfer \cite{Jaquier_2021_IJRR}. When such primitives are applied directly to curved geometries without geometric embedding, structural incompatibilities arise. Re-learning or re-synthesizing surface-intrinsic motion for each new geometry can be computationally expensive or data-intensive. What is missing is a principled embedding mechanism that resolves global incompatibility before execution while preserving structural properties such as periodicity and axis consistency.

Recent work has explored geometric policy transportation and diffeomorphic motion mappings to generalize manipulation skills across task configurations. Franzese et al.~\cite{Franzese_2025_TRO} learn nonlinear transportation maps that deform demonstrated policies between geometric domains through keypoint correspondences. Gao et al.~\cite{Gao2021MotionMappings_RAL} formulate invertible and continuously differentiable motion mappings for workspace adaptation in teleoperation. These approaches show that structured geometric transformations can preserve position, velocity, and orientation consistency during motion transfer. However, they primarily address policy transfer between task configurations rather than structured embedding of periodic motion primitives onto curved constraint manifolds under persistent contact.

Safety-aware execution methods, such as conic control barrier functions, bound orientation deviations within prescribed feasible sets during execution \cite{Shen_2024_ConicCBF, Tatsuya_2020_CSL}. These methods provide guarantees on bounded orientation error. However, they operate at the control layer and do not resolve geometric mismatch between a nominal motion primitive and a curved surface. Without an explicit embedding step, online tracking must continuously compensate for this mismatch. This increases corrective effort and may lead to drift over repeated cycles. The gap motivates a principled separation between offline geometric embedding and lightweight, contact-aware online projection.

\subsection{Our Contribution}

In this work, we decompose the problem into two structured stages. First, we introduce a \textbf{surface-constrained offline warping operator} that embeds a nominal periodic primitive onto a curved surface through diffeomorphic deformation with asymmetric track weighting and minimum-change orientation completion. This step resolves global geometric incompatibility while preserving internal structure such as periodicity and axis alignment. Second, we develop a \textbf{contact-aware online projection operator} that bounds deviation from the embedded reference through disturbance-triggered adjustment and a reference-centered conic orientation constraint, ensuring safe and drift-free execution during repeated surface interaction. By separating offline embedding from online projection, the framework enables stable surface-embedded execution under sensor-lite conditions without requiring prior surface-gripper calibration.

Our contributions are summarized as follows:
\begin{itemize}
    \item A surface-constrained offline warping formulation for embedding periodic motion primitives onto curved surfaces while preserving structural properties.
    \item A contact-aware online projection mechanism that bounds deviation and suppresses drift during repeated surface interaction.
    \item A two-stage execution framework that separates geometric embedding from online regulation, enabling safe operation with contact-sensor feedback and validation on real robotic experiments.
\end{itemize}

%% file: method.tex
\section{Methodology}
\label{sec:method}

\begin{figure}[!htb]
    \centering
    \includegraphics[width=0.48\textwidth]{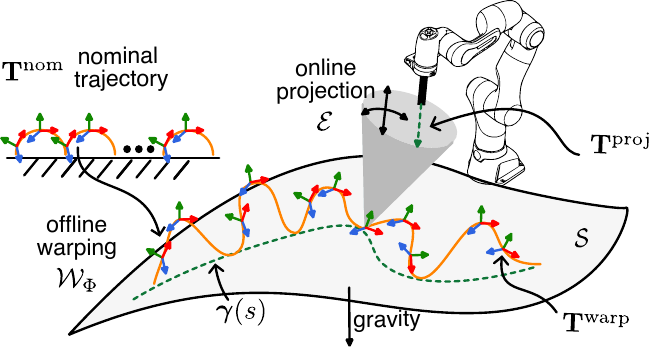}
    \caption{\textbf{Problem formulation and execution pipeline.}
A nominal periodic trajectory $\mathbf{T}^{\mathrm{nom}}$ is first adapted to a surface-embedded guide curve $\gamma(s)$ through the offline warping operator $\mathcal{W}_{\Phi}$, producing a surface-consistent warped trajectory $\mathbf{T}^{\mathrm{warp}}$ (orange). During execution, an online projection operator $\mathcal{E}$ constrains the executed pose $\mathbf{T}^{\mathrm{proj}}$ within a conic feasible set around the warped orientation while regulating position along gravity using contact-feedback measurements. This two-stage formulation separates offline geometric embedding from online safety-constrained execution.}
    \label{fig:problem_formulation}
\end{figure}

\begin{figure*}[!htb]
    \centering
    \includegraphics[width=0.9\textwidth]{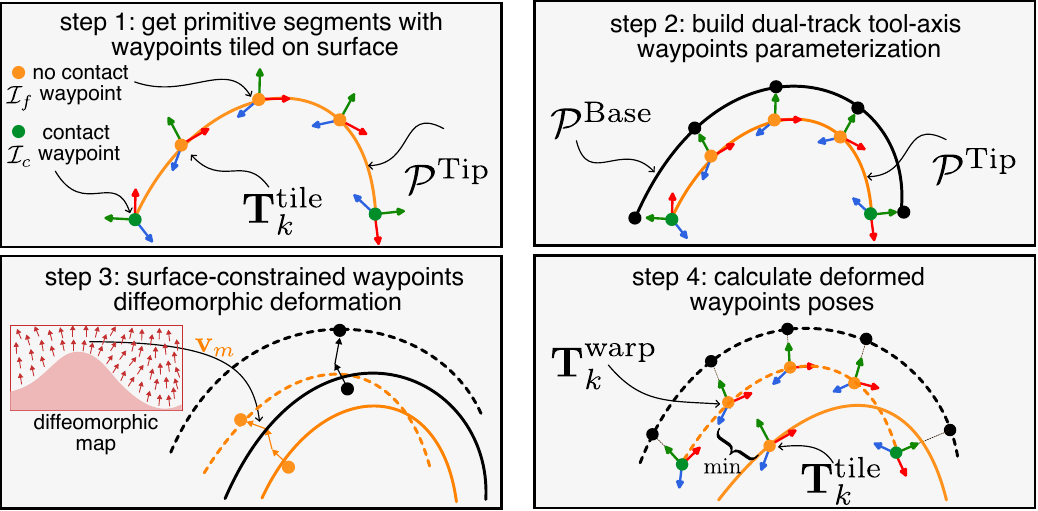}
    \caption{\textbf{Offline surface-constrained warping pipeline.} 
Step~1 extracts a single-period primitive and tiles its waypoints along a surface guide curve, separating contact ($\mathcal I_c$, green waypoints) and free-space ($\mathcal I_f$, orange waypoints) samples. 
Step~2 constructs a dual-track tool-axis parameterization using coupled base and tip waypoints. 
Step~3 applies a surface-constrained diffeomorphic deformation driven by local surface geometry.
Step~4 constructs the final warped pose sequence $\mathbf T_k^{\mathrm{warp}}$ by combining the deformed waypoints (with contact-critical waypoints anchored) with axis-consistent orientation completion.}
    \label{fig:offline_steps}
\end{figure*}

\subsection{Problem Formulation}

We assume an offline-defined surface-embedded guide curve $\boldsymbol{\gamma}(s) \subset \mathcal{S}$ representing the locus of desired contact points on a smooth surface $\mathcal{S}$, as shown in Fig.~\ref{fig:problem_formulation}. This curve encodes where periodic contact should occur and serves as the geometric backbone for trajectory generation.

Given $\boldsymbol{\gamma}(s)$, a remapped trajectory is constructed through the offline warping operator introduced in Section~\ref{sec:warping}. Starting from a predefined nominal periodic motion $\mathbf{T}^{\mathrm{nom}} = (\mathbf{R}^{\mathrm{nom}}, \mathbf{p}^{\mathrm{nom}}) \in SE(3)$, 
a surface-adapted pose sequence 
$\mathbf{T}^{\mathrm{warp}} = (\mathbf{R}^{\mathrm{warp}}, \mathbf{p}^{\mathrm{warp}}) \in SE(3) $
is generated via a surface-constrained warping of the nominal periodic motion along $\boldsymbol{\gamma}(s)$. This produces a periodic motion embedded on the surface (i.e., the orange trajectory in Fig.~\ref{fig:problem_formulation}).

During online projection, the executed pose 
$\mathbf{T}^{\mathrm{proj}} = (\mathbf{R}^{\mathrm{proj}}, \mathbf{p}^{\mathrm{proj}}) \in SE(3)$
is constrained such that its contact axis remains within a conic feasible set of half-angle $\theta$ centered at the warp direction,
\begin{equation}
\mathcal{C} =
\left\{
\mathbf{R}^{\mathrm{proj}} \in SO(3)
\;\middle|\;
\angle\!\left(
\mathbf{R}^{\mathrm{proj}}\mathbf{e}_c,\,
\mathbf{R}^{\mathrm{warp}}\mathbf{e}_c
\right)
\le \theta
\right\},
\end{equation}
where $\mathbf{e}_c \in \mathbb{R}^3$, $\|\mathbf{e}_c\| = 1$, denotes the canonical contact-direction unit vector expressed in the tool frame (e.g., the principal axis of a cylindrical end-effector). 
Although the cone is centered at the offline warped orientation $\mathbf{R}^{\mathrm{warp}}$, the constraint is enforced online to bound orientation deviation and prevent cumulative drift over repeated periodic embeddings. Meanwhile, the position is adjusted along the gravity direction $\hat{\mathbf g}$ using contact-feedback measurements (e.g., from a force-sensing resistor, FSR) to maintain persistent surface contact.

Together, the resulting framework consists of two operators: an offline warping operator $\mathcal{W}_{\Phi}$ and an online projection operator $\mathcal{E}$ that enforces conic safety constraints during execution,
\begin{equation}
\mathbf{T}^{\mathrm{warp}} = \mathcal{W}_{\Phi}\!\left(\mathbf{T}^{\mathrm{nom}}\right), \qquad
\mathbf{T}^{\mathrm{proj}} = \mathcal{E}\!\left(\mathbf{T}^{\mathrm{warp}}, F, \hat{\mathbf g}\right),
\label{eq:two_ops}
\end{equation}
where $\mathcal{W}_{\Phi}$ produces a surface-adapted trajectory from offline warping, and $\mathcal{E}$ generates the executed pose online under FSR-contact sensor reading $F$, along the gravity direction $\hat{\mathbf g}$.



\subsection{Offline Surface-Constrained Warping Operator $\mathcal{W}_{\Phi}$}
\label{sec:warping}

The offline surface-constrained warping operator $\mathcal W_\Phi$ transforms a tiled periodic primitive into a surface-consistent warped trajectory through a structured four-step procedure (Fig.~\ref{fig:offline_steps}). Specifically, we (i) extract a single-period primitive with associated waypoints and tile it along a surface guide curve, (ii) construct a dual-track tool-axis parameterization via tip and base waypoints, (iii) apply an asymmetric surface-constrained diffeomorphic deformation driven by local surface geometry to improve feasibility while preserving contact structure, and (iv) construct a smooth warped pose sequence via axis-consistent orientation completion. This staged formulation cleanly separates geometric embedding from pose construction and produces a stable surface-adapted trajectory for execution.

\subsubsection{Step 1: Primitive Segment Extraction and Surface-Based Tiling}

We begin with a contact-rich nominal periodic motion $\mathbf T^{\mathrm{nom}}$ and extract a single-period primitive segment together with its associated waypoints. Given a surface-embedded guide curve $\boldsymbol{\gamma}(s)\subset\mathcal S$ (or its discrete samples), we tile repeated instances of the primitive along the guide to obtain an initial surface-consistent placement (Fig.~\ref{fig:problem_formulation}).

In practice, we compute cumulative arc length along the guide and select approximately equal-distance tile centers within a prescribed tolerance. Each repeated period is aligned using the local chord direction between consecutive centers, producing an initial tiled pose sequence $\{\mathbf T_k^{\mathrm{tile}}\}_{k=1}^{K}$. From this sequence, we extract the tool-tip trajectory implied by the tiled poses and denote it as the tip-motion track $\mathcal P^{\mathrm{Tip}}=\{\mathbf p_k^{\mathrm{Tip}}\}_{k=1}^{K}$, which serves as the geometric carrier prior to refinement.

The tiled samples naturally divide into two waypoint types according to their contact role in the original demonstration: contact waypoints and non-contact waypoints. We encode this distinction using a contact-critical index set $\mathcal I_c\subset\{1,\ldots,K\}$ and a complementary free-space index set $\mathcal I_f=\{1,\ldots,K\}\setminus\mathcal I_c$, corresponding to the green (contact) and orange (non-contact) points shown in Step~2 of Fig.~\ref{fig:offline_steps}.

Although tiling captures the intended periodic placement along the guide, it does not explicitly account for local surface geometry. Consequently, poses determined solely by guide alignment may violate contact feasibility or exhibit non-smooth orientation variation in regions of high curvature or local concavity, motivating the deformation step described next.

\subsubsection{Step 2: Tool-Axis Parameterization via Dual-Track Waypoints}

Building on the tip-motion track obtained in Step~1, we represent the primitive using a tool-axis parameterization defined by two coupled waypoint trajectories: the tip track $\mathcal P^{\mathrm{Tip}}=\{\mathbf p_k^{\mathrm{Tip}}\}_{k=1}^{K}$ and the base track $\mathcal P^{\mathrm{Base}}=\{\mathbf p_k^{\mathrm{Base}}\}_{k=1}^{K}$ (Step~2 in Fig.~\ref{fig:offline_steps}). The base track is obtained directly from the translational component of the tiled poses, while the tip track represents the desired tool-tip motion extracted in Step~1. Together, these two tracks encode the primitive geometry in a geometry-aware manner.

At each sample $k$, the vector connecting the base and tip waypoints defines the instantaneous tool-axis direction after normalization. To ensure numerical stability, we require the baseline magnitude to exceed a small threshold; otherwise, the most recent valid axis direction is reused. This dual-track representation provides a compact and geometrically consistent description of the primitive that supports stable subsequent deformation.

\subsubsection{Step 3: Surface-Constrained Diffeomorphic Deformation}

To improve geometric feasibility while preserving contact structure, we apply a smooth deformation map $\Phi:\mathbb{R}^3\to\mathbb{R}^3$ to jointly remap the dual tracks into a surface-consistent configuration. The deformation is generated from the local surface geometry: specifically, the incremental deformation field is constructed using the surface normal direction at each spatial sample, so that corrections primarily act along the normal while preserving tangential structure. This encourages samples that violate contact feasibility to move toward the surface without introducing unnecessary lateral distortion.

The deformation is implemented as a composition of incremental updates,
\begin{equation}
\Phi=(\phi_1\circ\cdots\circ\phi_m),
\qquad
\phi_m(\mathbf p)=\mathbf p+\mathbf v_m(\mathbf p),
\end{equation}
where $\mathbf p\in\mathbb{R}^3$ denotes a spatial sample and $\mathbf v_m(\mathbf p)$ is a smooth normal-driven correction field derived from the local surface geometry at each update.

Rather than deforming both tracks from Step~2 identically, we apply track-dependent strengths so that the tip and base trajectories respond differently to geometric corrections,
\begin{equation}
\phi_m^{(j)}(\mathbf p)
=
\mathbf p+\lambda^{(j)}(\mathbf p)\,\mathbf v_m(\mathbf p),
\qquad j\in\{\mathrm{Tip},\mathrm{Base}\}.
\end{equation}
This asymmetric design improves both positional feasibility and orientation continuity under surface constraints: stronger corrections on the tip track enforce surface-consistent contact, while milder adjustments on the base track prevent excessive axis rotation and preserve smooth orientation evolution.

Applying the resulting diffeomorphic maps to both tracks from Step~2 yields the warped trajectories
\[
\tilde{\mathbf p}_k^{(j)}=\Phi^{(j)}(\mathbf p_k^{(j)}),
\qquad j\in\{\mathrm{Tip},\mathrm{Base}\},
\]
which serve as geometrically consistent inputs for reference pose construction in the next step. To preserve task semantics, contact-critical samples ($k\in\mathcal I_c$) remain anchored during deformation, while only samples in $\mathcal I_f$ are updated.

\subsubsection{Step 4: Pose Construction under $\Phi$}

Using the warped dual-track waypoints obtained in Step~3, we construct the final pose sequence under the deformation operator $\Phi$. The warped tip track directly defines the position sequence,
\begin{equation}
\label{eq:warped_position}
\mathbf p_k^{\text{warp}}=\tilde{\mathbf p}_k^{\mathrm{Tip}}
\end{equation}
The vector connecting the warped base and tip waypoints defines the tool-axis direction after normalization. If the baseline magnitude falls below a small threshold, we reuse the previous valid axis direction to ensure numerical stability.

Since the tool-axis direction (the green axes in Fig.~\ref{fig:offline_steps}) alone does not uniquely determine an element of $SO(3)$, one rotational degree of freedom remains unconstrained. To ensure smooth and consistent orientation evolution, we resolve this ambiguity using a minimum-change completion rule:
\begin{equation}
\mathbf R_k^{\text{warp}}
=
\arg\min_{\mathbf R\in SO(3)}
d(\mathbf R,\mathbf R_{k}^{\text{tile}})
\end{equation}
This selects, among all feasible orientations aligned with the tool axis, the one closest to the tiled pose, thereby suppressing unnecessary rotation and ensuring temporal smoothness. Together with the position in Eq.~\ref{eq:warped_position}, the resulting warped pose sequence is
\begin{equation}
\mathbf T_k^{\text{warp}}
=
\left(\mathbf R_k^{\text{warp}},\mathbf p_k^{\text{warp}}\right).
\end{equation}

Across Steps~1--4, the offline warping operator $\mathcal W_{\Phi}$ maps the nominal trajectory to the warped trajectory as
\begin{equation}
\mathbf T^{\text{warp}}=\mathcal W_{\Phi}\!\left(\mathbf T^{\text{nom}}\right).
\label{eq:two_ops}
\end{equation}


\subsection{Online Execution $\mathcal{E}$}
\label{sec:exec}

Given the offline warp pose
\(
\mathbf T_k^{\mathrm{warp}}
=
(\mathbf R_k^{\mathrm{warp}},\mathbf p_k^{\mathrm{warp}})
\in SE(3)
\),
the online operator \(\mathcal E\) generates the projected pose
\(
\mathbf T_{k}^{\mathrm{proj}}
=
(\mathbf R_{k}^{\mathrm{proj}},\mathbf p_{k}^{\mathrm{proj}})
\in SE(3)
\)
by combining FSR-driven local disturbance correction with an always-on conic safety filter:
\begin{equation}
\mathbf T_{k}^{\mathrm{proj}}
=
\mathcal E\!\left(
\mathbf T_k^{\mathrm{warp}},\, F_k,\, \hat{\mathbf g}
\right).
\label{eq:online_operator}
\end{equation}
\begin{figure}[!t]
    \centering
    \includegraphics[width=\linewidth]{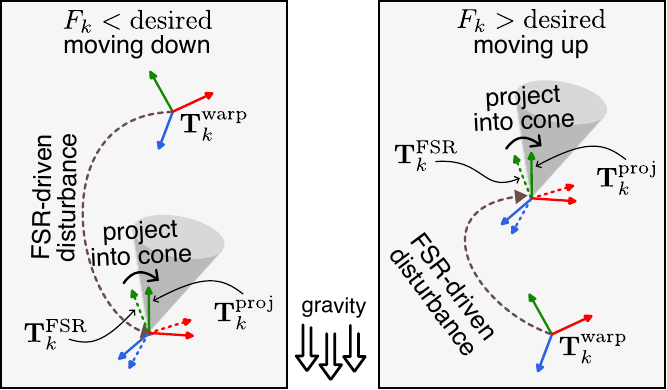}
    \caption{\textbf{Contact-aware online projection under FSR-driven disturbances.}
Force feedback perturbs the warped pose $\mathbf{T}_k^{\mathrm{warp}}$: low force drives downward motion, while high force drives upward motion. The disturbed pose is then projected into the conic feasible set, yielding $\mathbf{T}_k^{\mathrm{proj}}$ with bounded orientation deviation and stable contact.}
    \label{fig:online_pipeline}
\end{figure}

As illustrated in Fig.~\ref{fig:online_pipeline}, the online operator proceeds in two stages. First, FSR feedback generates a local candidate pose \((\mathbf R_k^{\mathrm{FSR}}, \mathbf p_k^{\mathrm{FSR}})\): insufficient force causes the tool to move downward and rotate toward gravity, whereas excessive force causes it to move upward and rotate away from gravity. Second, an always-on conic filter compares the candidate tool axis \(\mathbf u_k^{\mathrm{can}}\) with the warp axis \(\mathbf u_k^{\mathrm{warp}}\) and applies a bounded correction to produce the final projected orientation \(\mathbf R_k^{\mathrm{proj}}\).

\subsubsection{FSR-Driven Disturbance Candidate Generation}

At each step, we evaluate the contact error relative to the desired FSR level:
\begin{equation}
e_k = F_k - F^\star .
\label{eq:fsr_error}
\end{equation}
If \(|e_k|\le \varepsilon_F\), no disturbance correction is applied, i.e.,
\(
\Delta \mathbf p_k^{\mathrm{fsr}}=\mathbf 0
\)
and
\(
\Delta \mathbf R_k^{\mathrm{fsr}}=\mathbf I
\),
so that
\(
\mathbf T_k^{\mathrm{FSR}}=\mathbf T_k^{\mathrm{warp}}
\).

Otherwise, we generate bounded translational and rotational disturbance increments around the warp pose. The candidate translation is defined by
\begin{equation}
\mathbf p_k^{\mathrm{FSR}}
=
\mathbf p_k^{\mathrm{warp}}
+
\Delta_k\,\hat{\mathbf g},
\qquad
\Delta_k
=
\mathrm{sat}\!\big(\kappa_p e_k\big),
\label{eq:pcan_def}
\end{equation}
where \(\hat{\mathbf g}\) is the prescribed correction direction, chosen as the gravity direction in our implementation, \(\kappa_p\) is the translational gain, and
\(
\mathrm{sat}(\xi)=\min(\Delta_{\max},\max(-\Delta_{\max},\xi))
\)
bounds the correction magnitude.

Let
\(
\mathbf u_k^{\mathrm{warp}}=\mathbf R_k^{\mathrm{warp}}\mathbf e_c
\)
denote the warp tool axis, where \(\mathbf e_c\) is the designated tool-axis basis vector in the local tool frame. We define the FSR-induced rotation axis by
\begin{equation}
\mathbf a_k^{\mathrm{fsr}}
=
\frac{\mathbf u_k^{\mathrm{warp}}\times \hat{\mathbf g}}
{\|\mathbf u_k^{\mathrm{warp}}\times \hat{\mathbf g}\|+\epsilon},
\end{equation}
and compute the bounded disturbance angle
\begin{equation}
\delta_k^{\mathrm{fsr}}
=
\mathrm{sat}(\kappa_R e_k),
\qquad
|\delta_k^{\mathrm{fsr}}|
\le
\delta_{\max}^{\mathrm{fsr}}.
\end{equation}
The candidate orientation is then
\begin{equation}
\mathbf R_k^{\mathrm{FSR}}
=
\exp\!\big(
-[\delta_k^{\mathrm{fsr}}\mathbf a_k^{\mathrm{fsr}}]_\times
\big)\,
\mathbf R_k^{\mathrm{warp}}.
\label{eq:Rcan_def}
\end{equation}
Accordingly, the FSR-driven candidate pose is
\begin{equation}
\mathbf T_k^{\mathrm{FSR}}
=
(\mathbf R_k^{\mathrm{FSR}},\mathbf p_k^{\mathrm{FSR}}).
\end{equation}
By construction, the candidate reduces to the warp pose inside the deadband and becomes a bounded local disturbance otherwise.

\subsubsection{Always-On Conic Safety Constraints}

The conic filter is active at every step and acts only on orientation. Its role is to bound the deviation of the disturbed candidate orientation from the offline warp without removing the FSR-driven compliance entirely. Let
\(
\mathbf u_k^{\mathrm{can}}=\mathbf R_k^{\mathrm{FSR}}\mathbf e_c
\)
denote the candidate tool axis, and define the axis deviation angle by
\[
\phi_k
=
\angle(\mathbf u_k^{\mathrm{can}},\mathbf u_k^{\mathrm{warp}}).
\]
The admissible orientation set is given by a cone of half-angle \(\theta\) centered at \(\mathbf u_k^{\mathrm{warp}}\).

Whenever the candidate axis approaches or exceeds the cone boundary, the filter generates a bounded pushback rotation toward the warp axis. Using the correction axis
\begin{equation}
\mathbf a_k^{\mathrm{cone}}
=
\frac{\mathbf u_k^{\mathrm{can}}\times \mathbf u_k^{\mathrm{warp}}}
{\|\mathbf u_k^{\mathrm{can}}\times \mathbf u_k^{\mathrm{warp}}\|+\epsilon},
\end{equation}
the filtered orientation is computed as
\begin{equation}
\mathbf R_{k}^{\mathrm{proj}}
=
\exp\!\big(
[\delta_k^{\mathrm{cone}}\mathbf a_k^{\mathrm{cone}}]_\times
\big)\,
\mathbf R_k^{\mathrm{FSR}},
~
0 \le \delta_k^{\mathrm{cone}} \le \delta_{\max}^{\mathrm{cone}}.
\label{eq:R_out}
\end{equation}
Here \(\delta_k^{\mathrm{cone}}\) is a nonnegative correction magnitude that increases as the candidate approaches or exceeds the conic boundary.

Since the conic filter modifies only orientation, we set
\begin{equation}
\mathbf p_{k}^{\mathrm{proj}}=\mathbf p_k^{\mathrm{FSR}}.
\end{equation}
The final projected pose is therefore
\begin{equation}
\mathbf T_{k}^{\mathrm{proj}}
=
\left(
\mathbf R_{k}^{\mathrm{proj}},\,
\mathbf p_{k}^{\mathrm{proj}}
\right).
\label{eq:T_out}
\end{equation}
This composition preserves local translational contact adaptation while preventing excessive rotational drift away from the offline warp.